\documentclass[letterpaper]{article} % DO NOT CHANGE THIS
\usepackage{aaai24}  % DO NOT CHANGE THIS
\usepackage{times}  % DO NOT CHANGE THIS
\usepackage{helvet}  % DO NOT CHANGE THIS
\usepackage{courier}  % DO NOT CHANGE THIS
\usepackage[hyphens]{url}  % DO NOT CHANGE THIS
\usepackage{graphicx} % DO NOT CHANGE THIS
\usepackage{natbib}  % DO NOT CHANGE THIS AND DO NOT ADD ANY OPTIONS TO IT
\usepackage{caption} % DO NOT CHANGE THIS AND DO NOT ADD ANY OPTIONS TO IT
\usepackage{amsfonts}
\usepackage[frozencache=true,cachedir=minted-output]{minted}
\usepackage{algorithmic}
\usepackage{booktabs}
\usepackage{fontawesome}
\usepackage{subfig}
\usepackage{overpic}

\usepackage{multirow}
\usepackage{newfloat}
\usepackage{listings}
\DeclareCaptionStyle{ruled}{labelfont=normalfont,labelsep=colon,strut=off} % DO NOT CHANGE THIS
\floatstyle{ruled}
\newfloat{listing}{tb}{lst}{}
\floatname{listing}{Listing}
\title{Code-Style In-Context Learning for Knowledge-Based Question Answering}
\author{
    Zhijie Nie\textsuperscript{\rm 1,3},
    Richong Zhang\textsuperscript{\rm 1,2 \thanks{Corresponding author}},
    Zhongyuan Wang\textsuperscript{\rm 1},
    Xudong Liu\textsuperscript{\rm 1}
}
\affiliations{
    \textsuperscript{\rm 1}SKLSDE, School of Computer Science and Engineering, Beihang University, Beijing, China\\
    \textsuperscript{\rm 2}Zhongguancun Laboratory, Beijing, China\\
    \textsuperscript{\rm 3}Shen Yuan Honors College, Beihang University, Beijing, China \\
    \{niezj,zhangrc,wangzy23,liuxd\}@act.buaa.edu.cn
}

\usepackage{bibentry}
\begin{document}

\maketitle

\begin{abstract}
Current methods for Knowledge-Based Question Answering (KBQA) usually rely on complex training techniques and model frameworks, leading to many limitations in practical applications. Recently, the emergence of In-Context Learning (ICL) capabilities in Large Language Models (LLMs) provides a simple and training-free semantic parsing paradigm for KBQA: Given a small number of questions and their labeled logical forms as demo examples, LLMs can understand the task intent and generate the logic form for a new question. However, current powerful LLMs have little exposure to logic forms during pre-training, resulting in a high format error rate. To solve this problem, we propose a code-style in-context learning method for KBQA, which converts the generation process of unfamiliar logical form into the more familiar code generation process for LLMs. Experimental results on three mainstream datasets show that our method dramatically mitigated the formatting error problem in generating logic forms while realizing a new SOTA on WebQSP, GrailQA, and GraphQ under the few-shot setting. The code and supplementary files are released at \url{https://github.com/Arthurizijar/KB-Coder}.
\end{abstract}

\section{Introduction}
\begin{figure}[t]
    \centering
    \includegraphics[width=\linewidth]{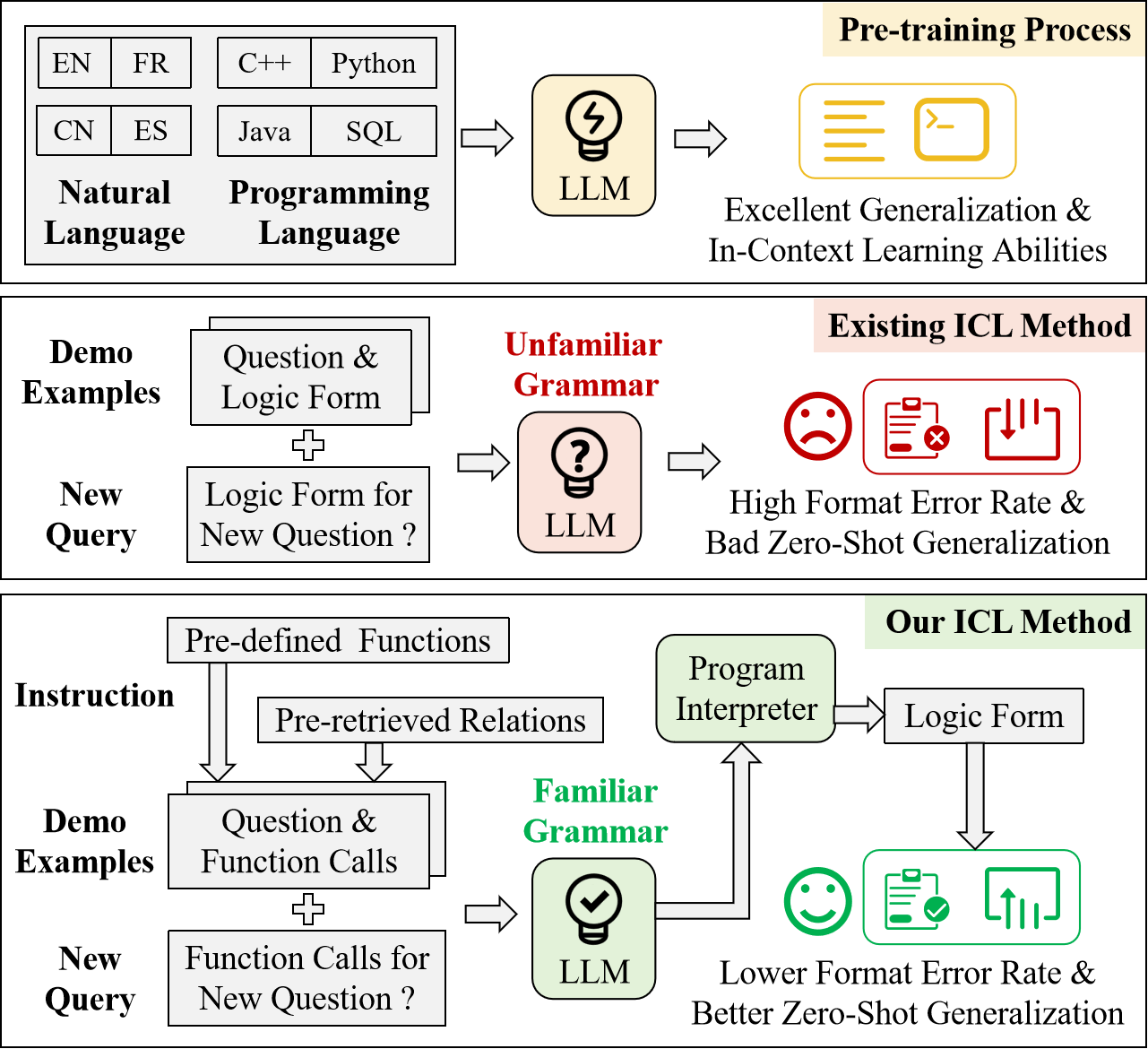}
    \caption{A comparison between our proposed ICL method and the existing method. Intuitively, our method achieves better performance by transforming the original KBQA task into a more familiar code form for the LLM.}
    \label{fig:introduction}
\end{figure}

Knowledge-Based Question Answering (KBQA) \cite{yih2016value,lan2020query,yu2022decaf,li2023shot} is a long-studied problem in the NLP community. Due to the complexity and diversity of the questions, the models with good performance usually have complex modeling frameworks or special training strategies. However, these designs also lead to models that require a lot of labeled data to help the parameters converge, making them difficult to apply in the new domains. Recently, Large Language Models (LLMs) have strong generalization capabilities, benefiting from pre-training on vast amounts of natural language corpus and open source code (Figure \ref{fig:introduction} Top). In addition, the emergence of In-Context Learning (ICL) capabilities \cite{brown2020language} allows LLMs to accomplish even complex reasoning tasks while observing a small amount of labeled data \cite{wei2022chain,cheng2022binding}.

Recently, Pangu \cite{gu2023dont} first proposed a KBQA method based on the ICL paradigm, which consists of a symbolic agent and a language model. The agent is responsible for constructing effective programs step-by-step, while the language model is responsible for guiding the search process to evaluate the plausibility of candidate plans. KB-BINDER \cite{li2023shot} proposes a training-free paradigm, which uses multiple (question, logical form) pairs as the demo examples, and leads the LLM to generate the correct logic form for the new question based on these demo examples (Figure \ref{fig:introduction} Middle). 
However, we find that this latest method still has several serious problems: (1) {\bf High format error rate}. Due to the highly specialized nature of the logic forms, all of them can hardly ever appear in the training corpus of LLM. As a result, it is challenging to generate the logical form with the correct formatting with a few demo examples. In our preliminary experiments, the logic forms generated by the powerful GPT3.5-turbo have a more than 15\% format error rate on GrailQA \cite{gu2021beyond}. Without the tedious syntactic parsing steps with error correction, the logic form generated cannot even provide a final answer to the question; (2) {\bf Low zero-shot generalization performance}. In the zero-shot generalization scenario, the prior knowledge of the KB domain schema related to the test question is unable to be obtained through the training set \cite{gu2021beyond}. In other words, it means that demo examples created from the training set cannot provide enough information for question answering under this scenario. Considering the labeled data usually covers only a small fraction of knowledge in KB, this problem will inevitably affect the performance in practice. To address these problems, the key is to deeply understand the behavior of the LLMs. Thankfully, some empirical observations on LLMs have provided desirable insights, which include (1) converting original tasks to code generation tasks will reduce the difficulty of post-processing \cite{wang2022code4struct}; (2) reasoning step-by-step can improve LLMs performance on complex reasoning tasks \cite{wei2022chain}; (3) retrieval augmentation is helpful for LLMs in dealing with uncommon factual knowledge \cite{mallen2023trust}.

Inspired by the above valuable observations, we propose a novel code-style in-context learning method for KBQA, which converts the one-step generation process of the logic forms to the progressive generation process of the function calls in Python (Figure \ref{fig:introduction} Bottom). Specifically, we first define seven meta-functions for S-Expression \cite{gu2021beyond}, a popular logic form for KBQA, and implement these functions in Python. Note that this step enables all S-Expressions to be generated by a finite number of calls of pre-defined meta-functions. For a test question, we sample a few numbers of (question, S-Expression) pairs in the training set and convert them into (question, function call sequence), where the function call sequence can be executed in the Python interpreter to output S-Expression. Finally, the Python implementation of the meta-functions, all (question, function call sequences) pairs, and the test question are reformatted in code form as input to the LLM. And the LLM is expected to complement a complete function call sequence for the new question to obtain the correct logic form. In addition, we find a simple and effective way to improve the performance in the zero-shot generalization scenario: Regard the test question as the query to retrieve a related relation in KB, and provide LLM with the relation for reference.

\begin{table*}[t]\scriptsize
    \centering
    \begin{tabular}{lll|l}
        \toprule
        {\bf Function} & {\bf Domain} & {\bf Range} & {\bf Mapping Descriptions} \\
        \midrule
        START & $\{E | E \in \mathcal{P}(\mathcal{E})\}$ & \multirow{4}{*}{$\{E^\prime | E \in \mathcal{P}(\mathcal{E})\}$} & Start from $E$ and return the same set $E^\prime = E$ \\
        % \midrule
        JOIN & $\{(r, E) | r \in \mathcal{R}, E \in \mathcal{P}(\mathcal{E})\}$ & & Join $r$ to $E$ and return $E^\prime$ pointed by $r$\\
        % \midrule
        AND & $\{(E_1, E_2) | E_1,E_2 \in \mathcal{P}(\mathcal{E})\}$ & & Returns the intersection $E^\prime$ of $E_1$ and $E_2$\\
        % \midrule
        CMP & $\{(c, r, v) | c \in \{{\rm <}, {\rm >}, {\rm \leq}, {\rm \geq}\}, r \in \mathcal{R}, v \in \mathcal{V})\}$ & & Returns the subset $E^\prime$ of $E$ whose value pointed by $r$ ${\rm <}$/${\rm >}$/${\rm \leq}$/${\rm \geq}$ than $v$ \\
        \midrule
        ARG & $\{(a, E, r) | a \in \{{\rm min}, {\rm max}\}, E \in \mathcal{P}(\mathcal{E}), r \in \mathcal{R}\}$ & $\{e | e \in \mathcal{E}\}$ & Select $e$ in $E$ whose value pointed by $r$ is largest / smallest \\
        \midrule
        COUNT & $\{E | E \in \mathcal{P}(\mathcal{E})\}$ & $\{n | n \in \mathbb{N}$\} & Return the element number $n$ of $E$ \\
        \midrule
        STOP & $\{E | E \in \mathcal{P}(\mathcal{E})\}$ & $-$ & Stop at $E$ and $E$ is regarded as the predicted answer set\\
        \bottomrule
    \end{tabular}
    \caption{Seven meta-functions with their domain, range, and mapping descriptions. $\mathcal{P}(.)$ represents the power set of a given set, $\mathcal{V}$ is a subset of $\mathcal{E}$ containing all entities in value type, and $\mathbb{N}$ represents the set of natural numbers.  For the other notations, please refer to the {\bf Overview} section.}
    \label{tab:meta_function}
\end{table*}

Our contribution can be summarized as follows:
\begin{itemize}
    \item We propose a novel code-style in-context learning method for KBQA. Compared to the existing methods, our method can effectively reduce the format error rate of the logic form generated by LLMs while providing additional intermediate steps during reasoning.
    \item We find that providing a question-related relation as a reference to LLMs in advance can effectively improve the performance in the zero-shot generalization scenario.
    \item We design a training-free KBQA model, KB-Coder, based on the proposed ICL method. Extensive experiments on WebQSP, GrailQA, and GraphQ show that our model achieves SOTA under the few-shot setting. While allowing access to the full training set, the training-free KB-Coder achieves competitive or better results compared with current supervised SOTA methods. 
\end{itemize}

\section{Related Work}

\paragraph{Complex Reasoning with Code-LLMs} 
Codex \cite{chen2021evaluating} first introduces a code corpus to train LLMs and finds that the obtained code-LLMs have excellent logical reasoning capabilities. Subsequently, code-LLMs have been used for a variety of complex tasks in two ways. The works in the first way only implicitly use the reasoning power that derives from code pre-training and proposes techniques such as chain-of-thought \cite{wei2022chain} and problem decomposition \cite{zhou2022least}, etc. And the works in another way convert the task form into code generation and guide the LLMs to achieve the original task goal by creating instances \cite{wang2022code4struct}, complementary code \cite{li2023codeie} or generating SQL \cite{cheng2022binding} or logic forms \cite{li2023shot} directly.

\paragraph{Question Answering with LLMs} 
We distinguish the different methods according to the type of LLM-generated content. The first class of methods guide LLMs to generate answer directly \cite{li2023chain,baek2023knowledge}. In these methods, the knowledge in the external knowledge source is converted into the index, then the questions are used as queries to obtain relevant knowledge from the index through sparse or dense retrieval. Then the questions and related knowledge are spliced together and fed into the LLM to generate the answers directly. The second class of methods \cite{izacard2022few,gu2023dont,tan2023make} views the LLM as a discriminator and leads LLMs to choose the correct answer or action from a candidate set. The third class of methods \cite{li2023shot,cheng2022binding} views the LLM as a semantic parser and guides the model to generate intermediate logic forms. Compared to generating the answer directly, the other two classes of methods can eliminate the risk of generating fake knowledge in principle but cannot achieve an end-to-end method. Beyond the above three classes, DECAF \cite{yu2022decaf} is a special case that directs LLM to generate both logic forms and the final answer by changing the prompt.

\section{Method}

\subsection{Overview}\label{sec:pre}
In general, our proposed KBQA method is a method based on semantic parsing: Given a natural language question $q$ and KB $\mathcal{G} = \{(h, r, t), h, t \in \mathcal{E}, r \in \mathcal{R}$\}, where $\mathcal{E}$ is the entity set and $\mathcal{R}$ is the relation set, our method can be viewed as a function $F$, which maps $q$ to a semantically consistent logic form $l = F(q)$. Then $l$ is converted into a query to execute, and the queried results are regarded as the answers to $q$. Specifically, we first design seven meta-functions, which can cover all atomic operations of a specific logic form, and re-define these meta-functions in Python. Finally, for a new question, the following three steps are adopted to get the answer (Figure \ref{fig:method}): (1) An LLM is adopted to obtain its function call sequence with the code-style in-context learning method; (2) A dense retriever is utilized to link entities for the entity mentions extracted from the function calls, while another dense retriever is utilized to match relations for the relation name extracted from the function calls; (3) A program interpreter is used to execute the generated function call sequence to get the logical form $l$, which will be executed further to get the answer $a$.

\subsection{Meta-Function Design}
\begin{figure}[ht]
    \centering
    \includegraphics[width=\linewidth]{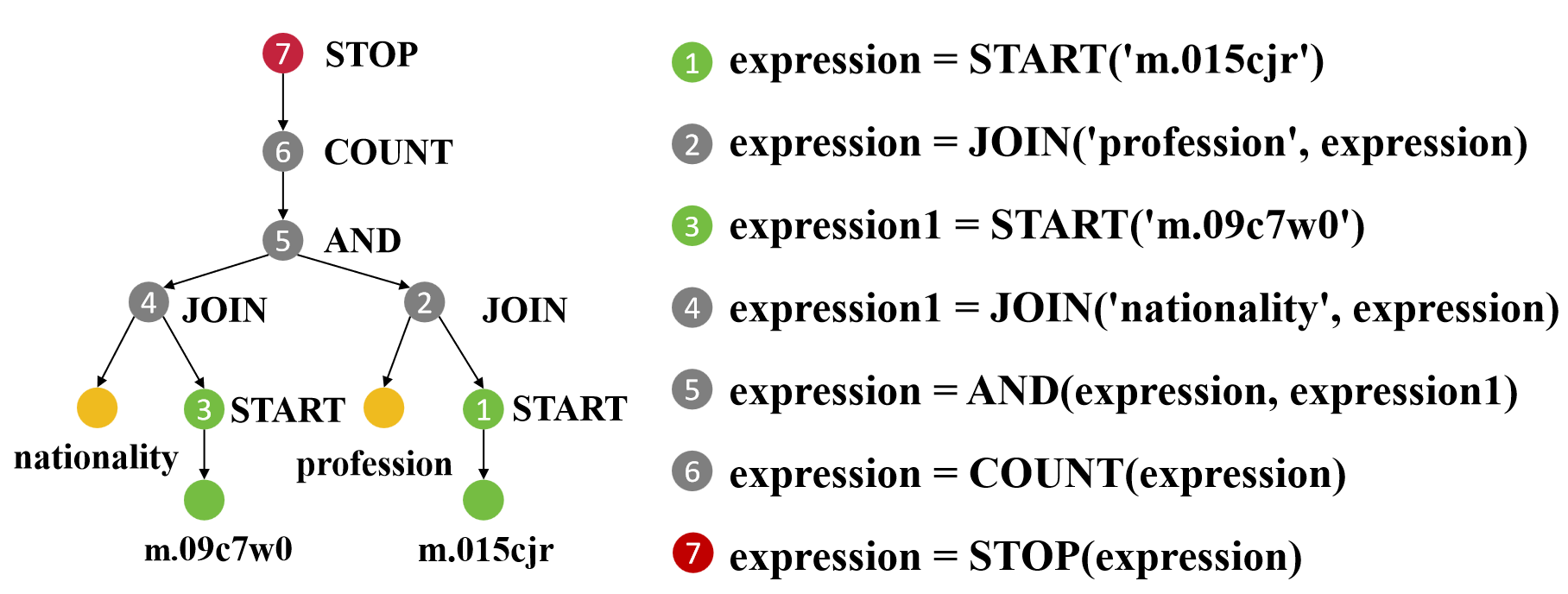}
    \caption{The tree structure (left) and the corresponding function call sequence (right) of S-Expression \texttt{(COUNT (AND (JOIN nationality m.09c7w0) (JOIN profession m.015cjr)))}.}
    \label{fig:tree_struct}
\end{figure}

In practice, we use S-Expression defined by \citet{gu2021beyond} as the logical form $l$ due to its simplicity. S-Expression \cite{mccarthy1960recursive} is a name-like notation for the nested list (tree-structured) data, which conforms to the grammar of ``prefix notation''. Specifically, S-Expression usually consists of parentheses and several space-separated elements within them, where the first element is the function name and all remaining elements are the attributes of the function. For example, for the question ``how many American presenters in total'', its S-Expression is

\begin{minted}{Python}
(COUNT (AND (JOIN nationality m.09c7w0)
            (JOIN profession m.015cjr)))
\end{minted}
where \texttt{m.09c7w0} and \texttt{m.015cjr} are the unique identifiers of entity \texttt{United States of America} and \texttt{Presenter}. The corresponding tree structure of this S-Expression is shown on the left side of Figure \ref{fig:tree_struct}.

Based on the original syntax of S-Expression \cite{gu2021beyond}, we define seven meta-functions in total (Table \ref{tab:meta_function}), which include their name, domain, range, and mapping description. Compared to the original grammar, we omit the {\bf R} function, remove a call way for the {\bf JOIN} function, and add the {\bf START} and {\bf STOP} functions.

\subsection{Code-Style In-Context Learning}
A typical in-context learning paradigm \cite{brown2020language} generally includes an instruction $I$, $K$ demo examples $D = \{d_1, d_2,...,d_K\}$, and a new query $Q$. If the output of the LLM is denoted as $C$, we can express the process of the in-context learning as
\begin{equation}
    C = f_{\rm LLM}(I;D;Q)
\end{equation}
where $f_{\rm LLM}$ represents a specific LLM. In our method, we use the code style to construct $I$, $D$, and $Q$ and expect the model to generate a piece of code - a sequence of meta-function calls - for $Q$, following the demo example. Next, we describe in detail how to construct each part respectively.

\begin{figure*}[t]
    \centering
    \includegraphics[width=0.9\linewidth]{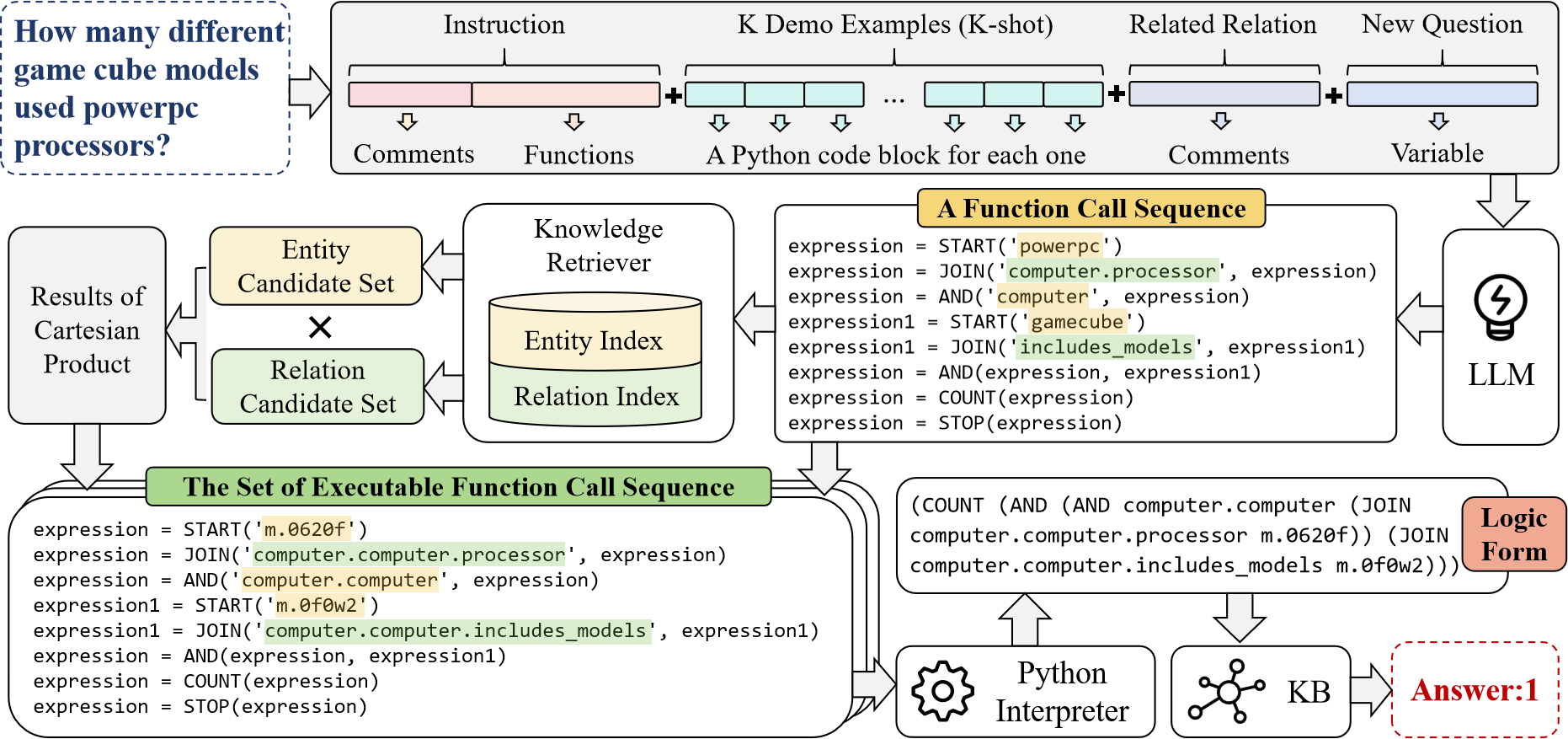}
    \caption{An illustration of the inference process of KB-Coder.}
    \label{fig:method}
\end{figure*}
\paragraph{Instruction $I$} Similar to previous works on code-LLMs \cite{cheng2022binding,li2023codeie}, the instruction consists of two parts: a prompt comment for describing the task and the code implementation of seven meta-functions. Due to the successful practice of Codex \cite{chen2021evaluating} in Python, we select Python to implement these functions. Then, the complete contents of $I$ are shown as follows:

% \vspace{1em}
\begin{scriptsize}
\begin{minted}{Python}
'''
Please use the functions defined below to generate the 
expression corresponding to the question step by step.
'''
def START(entity: str):
    return entity

def JOIN(relation:str, expression:str):
    return '(JOIN {} {})'.format(relation, expression)

def AND(expression:str, sub_expression:str):
    return '(AND {} {})'.format(expression, sub_expression)

def ARG(op:str, expression:str, relation:str):
    assert op in ['ARGMAX', 'ARGMIN']
    return '({} {} {})'.format(op, expression, relation)

def CMP(op:str, relation:str, expression:str):
    assert operator in ['<', '<=', '>', '>=']
    return '({} {} {})'.format(op, relation, expression)

def COUNT(expression:str):
    return '(COUNT {})'.format(expression)

def STOP(expression:str):
    return expression
\end{minted}
\end{scriptsize}
where all seven functions are the implementation of the same-name function in Table \ref{tab:meta_function}. 

\paragraph{Demo Examples $D$} Each demo example is created from a (question, S-Expression) pair in the training set and contains two parts: a variable named \texttt{question} and a function call sequence. Specifically, string variable \texttt{question} is assigned by the question, while the function call sequence is converted by the S-Expression. The right side of Figure \ref{fig:tree_struct} provides an example of the function call sequence, where each function call can correspond to one non-leaf node in the tree structure. Thus, the function call sequence can also be regarded as the result of a bottom-up parsing of the tree structure. If the meta-function definitions in Instruction are spliced together with the function call sequence as a whole code into the Python interpreter, we can get the complete S-Expression by visiting the value of variable \texttt{expression}. Since the entity identifiers in KB, such as \texttt{m.03yndmb}, lose their semantic information, we map all entity identifiers to their surface name to help the LLM obtain the semantic information of these entities. Finally, the demo example corresponding to Figure \ref{fig:tree_struct} is reformulated as follows:

\begin{scriptsize}
\begin{minted}{Python}
question = 'how many american presenters in total'
expression = START('presenter')
expression = JOIN('profession', expression)
expression1 = START('united States of America')
expression1 = JOIN('nationality', expression)
expression = AND(expression, expression1)
expression = COUNT(expression)
expression = STOP(expression)
\end{minted}    
\end{scriptsize}
To obtain $D$, we sample $K$ (question, S-Expression) in the training set in total, convert each pair to the above form, and split them using line breaks.

\paragraph{New Query $Q$} Compared to the form of demo examples, the new query is an incomplete piece of code and only contains the part of the variable \texttt{question}, For example, the question ``how many religious texts does syncretism have'' will be reformulated as
\begin{scriptsize}
\begin{minted}{Python}
question = 'how many religious texts does syncretism have'
\end{minted}    
\end{scriptsize}
For each $Q$, the LLM is expected to complement the remaining function call sequence corresponding to the question with the capability of in-context learning.

In practice, $I$ remains constant. For $D$, we can provide a consistent $D$ for each $Q$ in the dataset to follow the few-shot setting, or we can select a different $D$ for each $Q$ based on the similarity for better performance. Both two settings will be studied in subsequent experiments.

\subsection{Reasoning}
In this section, we describe the method to post-process the function call sequence generated by the LLM so that they can be converted into the query, ultimately obtaining the predicted answer.

\paragraph{Entity Linking \& Relation Matching}
Recall that the LLM can not have a direct view of the information in KB, making it difficult to generate completely correct entity names and relationship names. However, We believe that the names generated have a high level of semantic similarity with the correct ones, so the names generated can be treated as mentions for entity linking and relation matching. Benefiting from our strict definition of the domain of meta-functions, both entity mentions and relation mentions can be parsed easily. Specifically, for entity linking, we first convert all surface names of all entities in the KB into representations by the off-the-shelf embedding model, SimCSE \cite{gao2021simcse}, and build the entity index with Faiss \cite{johnson2019billion}. When linking for an entity mention, we first use the HARD MATCH strategy to obtain all entities that have the same surface name as the mention as the candidate set. If the size of the candidate set is larger than a hyper-parameter $M_{\rm e}$, we will only retain the most popular $M_e$ entities referring to FACC1 \cite{evgeniy2013facc1}. In contrast, if the size is less than $M_e$, we will search the most similar entities from the existing entity index to fill the candidates up to $M_{\rm e}$. Similarly, we use the same techniques to index all relations in the KB and retrieve $M_{\rm r}$ most semantically similar relations as candidates for each generated relation name. 
 
\paragraph{Answer Prediction}
Without loss of generality, we assume that the function call sequence contains $p$ entities to be linked and $q$ relations to be matched, denote the candidate set of the i-th entity mention as $\mathbb{C}_{\rm e}^i$ and the candidate set of the j-th relation mention as $\mathbb{C}_{\rm r}^j$. Then we can obtain the ordered tuple set $\{(c_{\rm e}^1,...,c_{\rm e}^p, c_{\rm r}^1, ..., c_{\rm r}^q)\} = \mathbb{C}_{\rm e}^1\times...\times\mathbb{C}_{\rm e}^p\times\mathbb{C}_{\rm r}^1\times...\times\mathbb{C}_{\rm r}^q$, where $\times$ represents the Cartesian product. For each ordered tuple in the set, let each element in the ordered tuple replace the corresponding generated names in the function call sequence one by one, we can obtain a candidate for the whole function call sequence (Figure \ref{fig:method}). Finally, there will be $(M_{e})^p.(M_{r})^q$ candidate items for each function call sequence, which can be a huge number for some special cases. Therefore, we will execute the sequence of function calls one by one to get the S-Expression, and then convert the S-Expression to SPARQL to execute it. Once the queried result is not null, we will just terminate the process of trying one by one and consider the result of the queried result as the answer to the question.

\subsection{One Related Relation for Zero-shot Generalization}
Code-Style in-context learning allows the LLM to be more adaptable to the task form, resulting in a lower format error rate. However, when the queried domain is not involved in the demo examples, which is called the zero-shot generalization scenario by \cite{gu2021beyond}, the performance of the LLM is still poor. This is not difficult to understand, as the change in the prompt form of the task does not bring new additional knowledge of the queried domain to the LLM. After analyzing the bad cases from the preliminary experiments, we found that the biggest problem with the error was relation matching, where the relation mentions generated by the LLM usually cannot hit the correct relation under the zero-shot generalization scenario. Subsequently, we mitigate this problem by providing an additional relation name for LLM. Specifically, we use the entire test question as a query to retrieve the similar relation from the relation index, and the relation with the highest similarity is inserted between demo examples $D$ and the new question $Q$ with the following comment format:
\begin{scriptsize}
\begin{minted}{Python}
'''
Some relations for reference are as follows:
conferences.conference_sponsor.conferences
'''
\end{minted}
\end{scriptsize}
Based on preliminary experiments, it is observed that one relation brings the best performance and more relations have little improvement on the performance. We will analyze the effect of the number of relations on the results in detail in the {\bf Experiment} section.

\section{Experiment}
\subsection{Experiment Setup}

\paragraph{Dataset} We use three mainstream datasets in KBQA, WebQSP \cite{yih2016value}, GraphQ \cite{su2016generating}, and GrailQA \cite{gu2021beyond}, which represent the three generalization capabilities of i.i.d, compositional, and zero-shot, respectively, to evaluate the effect of KB-Coder.

\paragraph{LLM} Due to the deprecation of the Codex family of models, we select \texttt{gpt-3.5-turbo} from OpenAI for our experiments. In all experiments, we used the official API \footnote{\url{https://openai.com/api}} to obtain model results, where \texttt{temperature} is set to 0.7, \texttt{max\_tokens} is set to 300, and other parameters are kept at default values.

\begin{table*}[t]\footnotesize
    \centering
    \renewcommand\arraystretch{0.8}
    \begin{tabular}{lcccccc|ccc}
        \toprule
        \multirow{2}{*}{\bf Method} & \multicolumn{2}{c}{\bf I.I.D} & \multicolumn{2}{c}{\bf Compositional} & \multicolumn{2}{c}{\bf Zero-Shot} \vline & \multicolumn{3}{c}{\bf Overall}\\
        \cmidrule(l){2-3}\cmidrule(l){4-5}\cmidrule(l){6-7}\cmidrule(l){8-10}
        & {\bf EM} & {\bf F1} & {\bf EM} & {\bf F1} & {\bf EM} & {\bf F1} & {\bf FER}$\downarrow$ & {\bf EM} & {\bf F1} \\
        \midrule
        \multicolumn{10}{c}{\it Full Supervised on the Entire Training Set} \\
        \midrule
        RnG-KBQA \cite{ye2022rng} & 86.7 & 89.0 & 61.7 & 68.9 & 68.8 & 74.7 & - & 69.5 & 76.9\\
        DecAF \cite{yu2022decaf} & 88.7 & 92.4 & 71.5 & 79.8 & 65.9 & 77.3 & - & 72.5 & 81.4 \\
        TIARA \cite{shu2022tiara} & 88.4 & 91.2 & 66.4 & 74.8 & 73.3 & 80.7  & - & 75.3 & 81.9 \\
        \midrule
        \multicolumn{10}{c}{\it In-Context Learning (Training-Free)} \\
        \midrule
        KB-BINDER (1) & 40.0(2.3) & 43.3(2.7) & 33.9(2.7) & 36.6(2.6) & 40.1(3.6) & 44.0(4.1) & 20.0(2.4) & 38.7(3.0) & 42.2(3.3) \\
        KB-Coder (1) & \bf 40.6(3.3) & \bf 45.5(2.8) & \bf 34.5(3.6) & \bf 38.6(3.5) & \bf 42.2(5.9) & \bf 47.3(5.4) & \bf 3.0(0.9) & \bf 40.1(3.7) & \bf 44.9(3.4) \\
        \midrule
        KB-BINDER (6) & 43.6(2.1) & 48.3(2.5) & \bf 44.5(2.3) & 48.8(2.7) & 37.5(2.4) & 41.8(2.8) & 8.1(0.4) & 45.7(2.3) & 50.8(2.8) \\
        KB-Coder (6) & \bf 43.6(3.7) & \bf 49.3(3.3) & 44.0(2.2) & \bf 49.6(1.9) & \bf 37.7(2.6) & \bf 43.2(2.6) & \bf 0.6(0.2) & \bf 45.9(3.9) & \bf 51.7(3.3) \\
        \midrule
        KB-BINDER (1)-R & 74.7(0.1) & 79.7(0.1) & 44.6(0.4) & 48.5(0.5) & 37.1(0.2) & 40.8(0.1) & 16.4(0.2) & 47.6(0.0) & 51.7(0.1) \\
        KB-Coder (1)-R & \bf 76.2(3.0)  & \bf 80.2(1.9)  & \bf 50.4(0.7)  & \bf 54.8(0.7)  & \bf 45.8(0.4)  & \bf 50.6(0.9) & \bf 3.1(0.4)  & \bf 54.0(1.0)  & \bf 58.5(1.0)  \\
        \midrule
        KB-BINDER (6)-R & 75.8(0.1) & 80.9(0.1) & 48.3(0.4) & 53.6(0.4) & 45.4(0.2) & 50.7(0.3) & 5.2(0.1) & 53.2(0.1) & 58.5(0.1) \\
        KB-Coder (6)-R & \bf 76.9(3.1)  & \bf 81.0(1.8)  & \bf 52.7(0.9)  & \bf 57.8(1.0)  & \bf 48.9(0.2)  & \bf 54.1(0.6) & \bf 1.5(0.1)  & \bf 56.3(0.9)  & \bf 61.3(1.0)  \\
        \bottomrule
    \end{tabular}
    \caption{40-shot results on the local dev set of GrailQA. The values in parentheses indicate standard deviation.}
    \label{tab:grail_results}
\end{table*}

\begin{table}[ht]\small
    \centering
    \renewcommand\arraystretch{0.8}
    \begin{tabular}{lcc}
        \toprule
        {\bf Method} & {\bf FER}$\downarrow$ & {\bf F1} \\
        \midrule
        \multicolumn{3}{c}{\it Full Supervised on the Entire Training Set} \\
        \midrule
        ArcaneQA \cite{gu2022arcaneqa} & - & 75.6 \\
        TIARA \cite{shu2022tiara} & - & 76.7 \\
        DecAF \cite{yu2022decaf} & - & 78.7 \\
        \midrule
        \multicolumn{3}{c}{\it In-Context Learning (Training-Free)} \\
        \midrule
        KB-BINDER (1) & 3.9(1.1) & 52.6(1.1) \\
        KB-Coder (1) & \bf 1.9(2.3) & \bf 55.7(1.3) \\
        \midrule
        KB-BINDER (6) & 0.3(0.4) & 56.6(1.7) \\
        KB-Coder (6) & \bf 0.1(0.1) & \bf 60.5(1.9)\\
        \midrule
        KB-BINDER (1)-R & 1.9(0.3) & 68.9(0.3) \\
        KB-Coder (1)-R &  \bf 1.7(0.3) & \bf 72.2(0.2) \\
        \midrule
        KB-BINDER (6)-R & 0.7(0.0) & 71.1(0.2) \\
        KB-Coder (6)-R & \bf 0.3(0.2) & \bf 75.2(0.5) \\
        \bottomrule
    \end{tabular}
    \caption{100-shot results on WebQSP. The values in parentheses indicate standard deviation.}
    \label{tab:wqsp_results}
\end{table}

\begin{table}[ht]\small
    \centering
    \renewcommand\arraystretch{0.8}
    \begin{tabular}{lcc}
        \toprule
        {\bf Method} & {\bf FER}$\downarrow$ & {\bf F1}\\
        \midrule
        \multicolumn{3}{c}{\it Full Supervised on the Entire Training Set} \\
        \midrule
        SPARQA \cite{sun2020sparqa} & - & 21.5 \\
        BERT + Ranking \cite{gu2021beyond} & - & 25.0 \\
        ArcaneQA \cite{gu2022arcaneqa} & - & 31.8 \\
        \midrule
        \multicolumn{3}{c}{\it In-Context-Learning (Training-Free)} \\
        \midrule
        KB-BINDER (1) & 15.4(1.6) & 27.1(0.5) \\
        KB-Coder (1) & \bf 6.3(2.5) & \bf 31.1(1.3) \\
        \midrule
        KB-BINDER (6) & 2.8(0.6) & 34.5(0.8) \\
        KB-Coder (6) & \bf 0.6(0.2) & \bf 35.8(0.6) \\
        \midrule
        KB-BINDER (1)-R & 19.1(0.1) & 26.7(0.3) \\
        KB-Coder (1)-R & \bf 10.0(0.2) & \bf 30.0(0.3) \\
        \midrule
        KB-BINDER (6)-R & 5.7(0.5) & 32.5(0.5) \\
        KB-Coder (6)-R & \bf 0.9(0.1) & \bf 36.6(0.2) \\
        \bottomrule
    \end{tabular}
    \caption{100-shot results on GraphQ. The values in parentheses indicate standard deviation.}
    \label{tab:graph_results}
\end{table}

\paragraph{Baseline} We mainly compare our model with KB-BINDER \cite{li2023shot}, the SOTA model on WebQSP, GrailQA, and GraphQ under the few-shot setting. The original results of KB-BINDER are obtained with the deprecated \texttt{code-davinci-002}, and we reproduce their method with \texttt{gpt-3.5-turbo} while the other setting remains consistent with us for a fair comparison. Some results obtained by training on the whole training dataset \cite{ye2022rng,shu2022tiara,gu2021beyond,sun2019pullnet} are also reported for reference. 

\paragraph{Evaluation Metric} Consistent with previous works \cite{yu2022decaf,li2023chain}, we report F1 Score on WebQSP and GraphQ, while Exact Match (EM) and F1 Score on GrailQA as performance metrics. At the same time, we use the Format Error Rate (FER) to evaluate the proportion of logical forms generated by different methods that conform to the grammar of S-Expresssion.

\subsection{Implementation details}
Without special instructions, we reported the experiment results with $M_{\rm e}=15$ and $M_{\rm r}=100$. SimCSE instance \texttt{sup-simcse-bert-base-uncased}\footnote{\url{https://huggingface.co/princeton-nlp/sup-simcse-bert-base-uncased}} is used to obtain the dense representations. Consistent with KB-BINDER \cite{li2023shot}, we conduct 100-shot for WebQSP and GraphQ, and 40-shot for GrailQA. In the following sections, we use different notations to represent different variants:
\begin{itemize}
    \item {\bf KB-Coder (K)}: The fixed questions sampled randomly from the training set are selected to be the demo examples. The LLM generates K candidates and uses the majority vote strategy \cite{wang2022self} to select the final answer. The performance is expected to be further improved thanks to the self-consistency of the LLM.
    \item {\bf KB-Coder (K)-R}: Compared to KB-Coder (K), the questions most similar to every test question are selected as their demo examples.
\end{itemize}
Note that {\bf KB-Coder (K)} is strictly under the {\bf few-shot} setting, while {\bf KB-Coder (K)-R} is to explore the upper bound on performance when the whole training set can be accessed. We report results for K = 1 and 6 for fair comparison with KB-BINDER. For each setting, we rerun it three times and report the mean and standard deviation.
\begin{figure*}[th]
    \centering
    \subfloat[Effect of the shot number]{\includegraphics[width=0.33\linewidth]{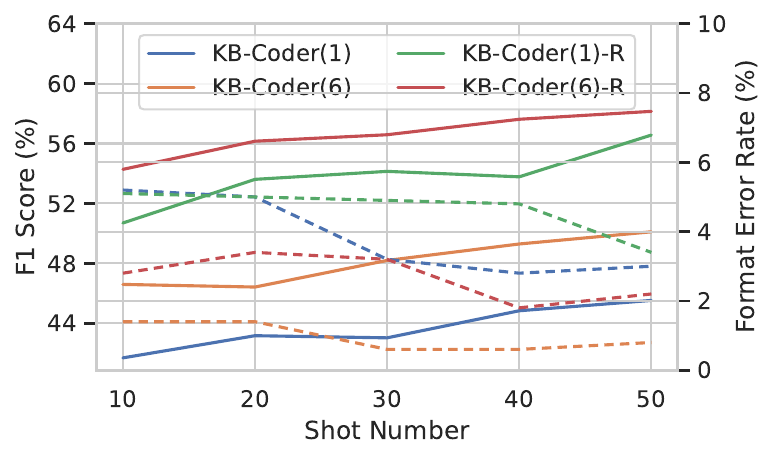}\label{fig:shot_number}}
    \subfloat[Effect of the relation number]{\includegraphics[width=0.33\linewidth]{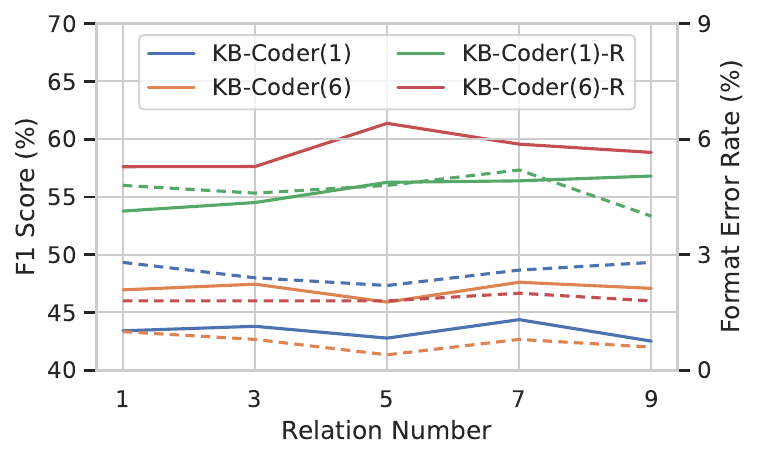}\label{fig:rel_number}}
    \subfloat[Effect of the vote number]{\includegraphics[width=0.33\linewidth]{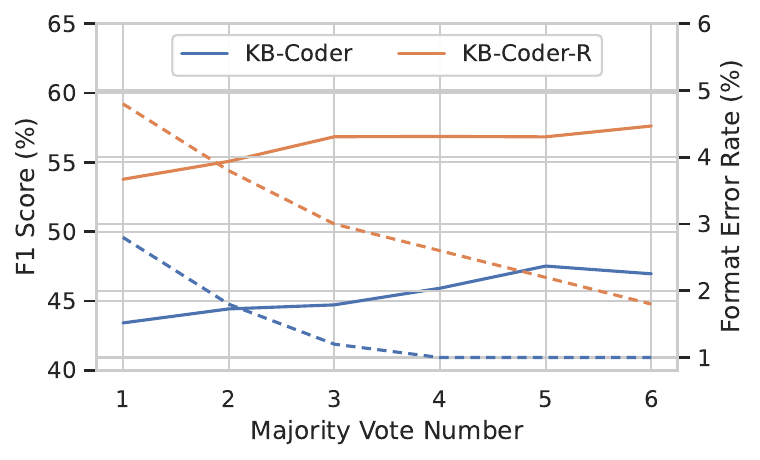}\label{fig:self_consist}}
    \caption{Effect analysis of the three factors in ICL with a subset of 500 questions from GrailQA local dev set, where the solid line is used to indicate F1 Score and the dashed line is used to indicate FER.}
\end{figure*}

\subsection{Main Results}
We report the performance of KB-Coder and other baselines on GrailQA, WebQSP, and GraphQ in Tables \ref{tab:grail_results}, \ref{tab:wqsp_results} and \ref{tab:graph_results}. Next, we will analyze the format error rate (FER) and the performance metrics (EM and F1) respectively.

Recall that one of the motivations for converting logical form generation to code generation is to allow LLM to do a more familiar task to ensure the correctness of the format of the generated content. the performance on all three datasets successfully verifies the effectiveness of our method under the few-shot setting: (1) On WebQSP, where the questions are simpler, the existing method generates logical forms with a low FMR, but KB-Coder can still further reduce the FMR to even lower levels; (2) On GrailQA and GraphQ, where the questions are more complex, KB-Coder improves dramatically compared to the two methods generating logic form directly; (3) Benefiting from self-consistency in LLMs, the majority vote strategy will help alleviate the problem of KB-BINDER format error rates. While KB-Coder can work with the majority vote strategy to promote new lows in FMR.

Benefiting from the lower FER, the F1 Scores (or EM) on the three datasets are both significantly improved compared to KB-BINDER under almost all settings, especially with two settings that do not introduce the majority vote strategy. (1) On WebQSP and GraphQ, the training-free KB-Coder(6)-R achieves competitive results compared to full-supervised methods, while our method further narrows down the performance with those of the fully-supervised model On GrailQA; (2) Compared to KB-BINDER, KB-Coder usually obtains a substantial lead under no dependence, reflecting the better underlying performance of our method; (3) The performance fluctuations of KB-Coder are more drastic compared to KB-BINDER, which is an issue we should focus to solve in the future work.

\subsection{Ablation Study}
\begin{figure}[th]
    \centering
    \subfloat[The ablation study on F1 Score.]{\includegraphics[width=\linewidth]{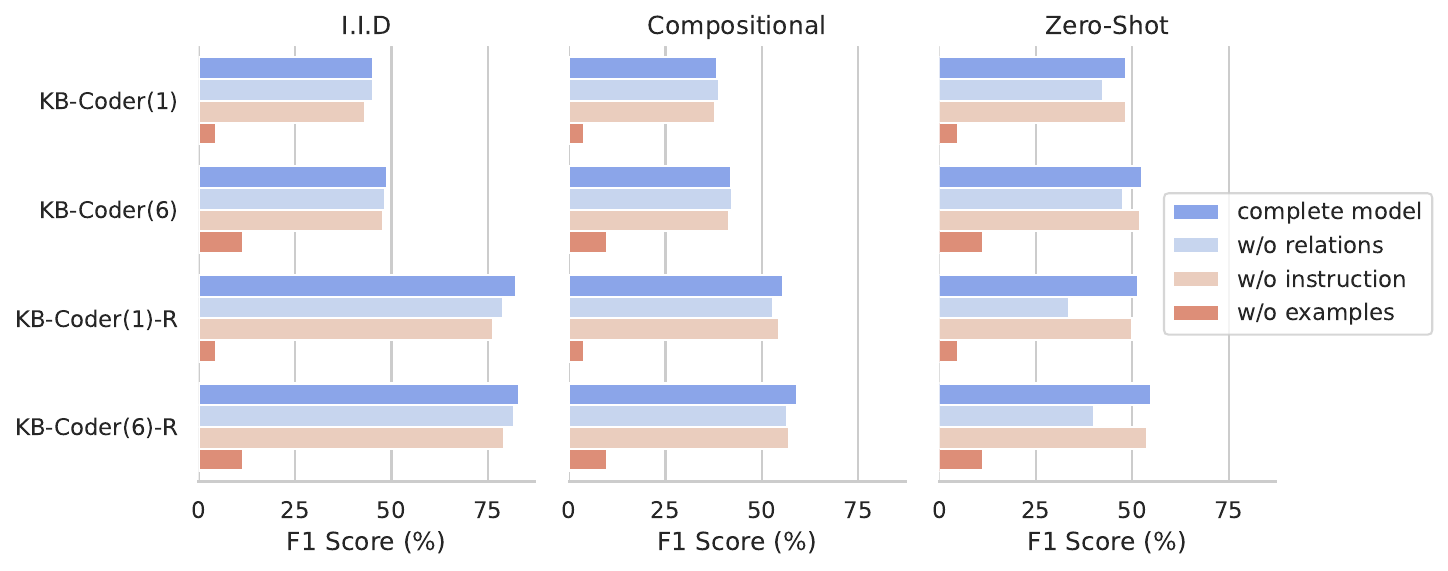}} \\
    \subfloat[The ablation study on Format Error Rate.]{\includegraphics[width=\linewidth]{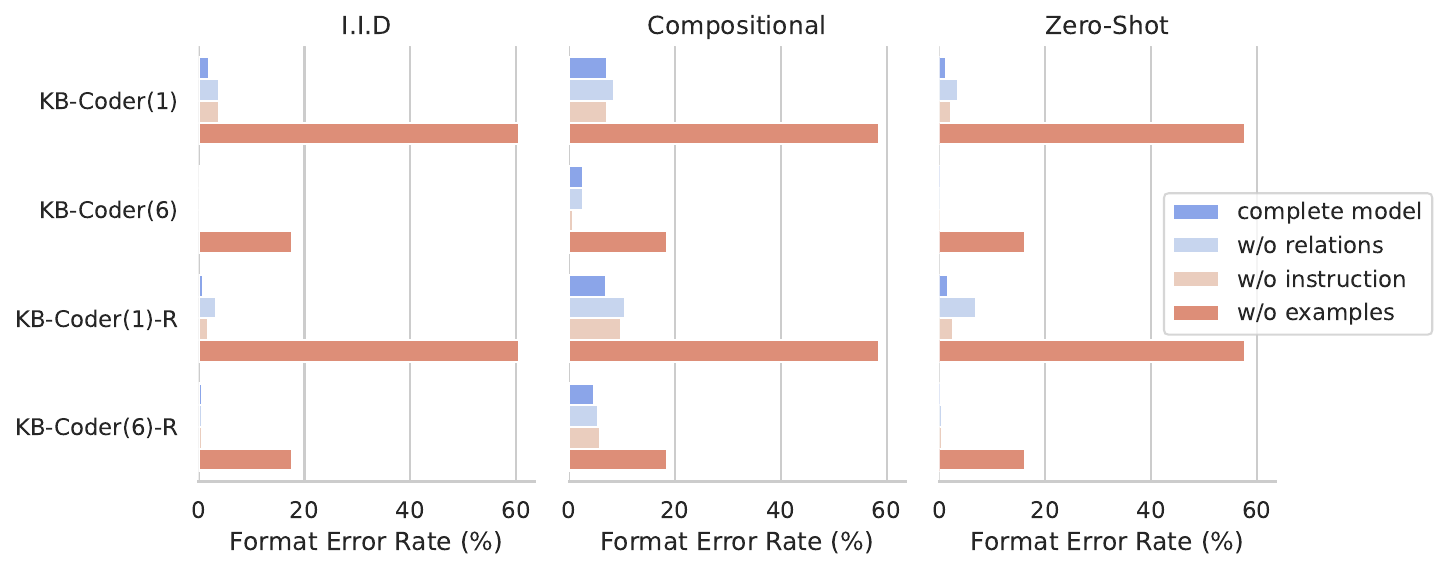}}\\
    \caption{Ablation Study on GrailQA.}
    \label{fig:similar_sampling}
\end{figure}

To explore the necessity of all parts of our method, we consider three settings in the ablation experiments: (1) removing related relation (-w/o relations) (2) removing instructions (-w/o instruction); (3) removing demo examples (-w/o examples). We report the results of three generalization levels in GrailQA separately in Figure \ref{fig:similar_sampling}. From the results, it can be seen that removing relations drastically reduces the F1 performance of the questions for zero-shot generalization, while having little impact on i.i.d and compositional questions. On top of that, removing instruction would bring about a weak degeneration in the F1 and FER. And removing demo examples would render the entire paradigm nearly invalid, meaning that it would be difficult for LLMs to understand the task requirements based on instructions alone.

\subsection{Analysis on In-Context Learning}
In this section, we explore the impact of three factors in ICL: the number of demo examples, the number of related relations, and the number of participating majority votes on the performance. For cost reasons, we performed this experiment on a subset of the 500-size scale on GrailQA.
\paragraph{The Effects of Demo Examples} We analyze the effect of the shot number on the results. Specifically, we set $K=\{10,20,30,40,50\}$ and report the performance trend with the shot number in Figure \ref{fig:shot_number}. The results show that boosting the number of shot numbers has a stabilizing effect on F1, whereas FER was maintained at a low level throughout.

\paragraph{The Effects of Related Relation Number} Similarly, we explore the effect of the number of correlations on the results. We set the related relation number as $\{1,3,5,7,9\}$ respectively and report the performance trend in Figure \ref{fig:rel_number}. The results show that the introduction of more relations has little effect on both F1 and FER. Instead, too many relations make the performance degrade.

\paragraph{The Effects of Answer Number}  We explore the effect of the answer number of participating in the majority vote on the performance. Specifically, we set the answer number as $\{1,2,3,4,5,6\}$ respectively and report the performance trend in Figure \ref{fig:self_consist}. The results show that FER improves significantly as the number of results participating in the vote rises, and there was an upward trend in F1 as well.

\section{Conclusion}
In this paper, we design a training-free KBQA framework, KB-Coder, which centers on a code-style in-context learning method for reducing formatting errors in generated logic forms and a retrieval-augment method for boosting zero-shot generalization capability. Extensive experimental results demonstrate that variants of our model, KB-Coder(1) and KB-Coder(6), achieve the SOTA performance under the few-shot setting, while the other two variants, KB-Coder(1)-R and KB-Coder(6)-R, achieves competitive performance compared to fully-supervised methods in a training-free premise. In general, KB-Coder demonstrates the potential of code-style ICL in KBQA and offers a training-free but effective baseline for the community.

\section{Acknowledgments}
This work was supported by the National Key R\&D Program of China under Grant 2022ZD0120200, in part by the National Natural Science Foundation of China (No. U23B2056), in part by the Fundamental Research Funds for the Central Universities, and in part by the State Key Laboratory of Software Development Environment.

\bibliography{aaai24}

\begin{thebibliography}{29}
\providecommand{\natexlab}[1]{#1}

\bibitem[{Baek, Aji, and Saffari(2023)}]{baek2023knowledge}
Baek, J.; Aji, A.~F.; and Saffari, A. 2023.
\newblock Knowledge-Augmented Language Model Prompting for Zero-Shot Knowledge Graph Question Answering.
\newblock \emph{arXiv preprint arXiv:2306.04136}.

\bibitem[{Brown et~al.(2020)Brown, Mann, Ryder, Subbiah, Kaplan, Dhariwal, Neelakantan, Shyam, Sastry, Askell et~al.}]{brown2020language}
Brown, T.; Mann, B.; Ryder, N.; Subbiah, M.; Kaplan, J.~D.; Dhariwal, P.; Neelakantan, A.; Shyam, P.; Sastry, G.; Askell, A.; et~al. 2020.
\newblock Language models are few-shot learners.
\newblock \emph{Advances in neural information processing systems}, 33: 1877--1901.

\bibitem[{Chen et~al.(2021)Chen, Tworek, Jun, Yuan, Pinto, Kaplan, Edwards, Burda, Joseph, Brockman et~al.}]{chen2021evaluating}
Chen, M.; Tworek, J.; Jun, H.; Yuan, Q.; Pinto, H. P. d.~O.; Kaplan, J.; Edwards, H.; Burda, Y.; Joseph, N.; Brockman, G.; et~al. 2021.
\newblock Evaluating large language models trained on code.
\newblock \emph{arXiv preprint arXiv:2107.03374}.

\bibitem[{Cheng et~al.(2022)Cheng, Xie, Shi, Li, Nadkarni, Hu, Xiong, Radev, Ostendorf, Zettlemoyer et~al.}]{cheng2022binding}
Cheng, Z.; Xie, T.; Shi, P.; Li, C.; Nadkarni, R.; Hu, Y.; Xiong, C.; Radev, D.; Ostendorf, M.; Zettlemoyer, L.; et~al. 2022.
\newblock Binding language models in symbolic languages.
\newblock \emph{arXiv preprint arXiv:2210.02875}.

\bibitem[{Gabrilovich, Ringgaard, and Subramanya(2013)}]{evgeniy2013facc1}
Gabrilovich, E.; Ringgaard, M.; and Subramanya, A. 2013.
\newblock FACC1: Freebase annotation of ClueWeb corpora, Version 1 (Release date 2013-06-26, Format version 1, Correction level 0).

\bibitem[{Gao, Yao, and Chen(2021)}]{gao2021simcse}
Gao, T.; Yao, X.; and Chen, D. 2021.
\newblock {SimCSE}: Simple Contrastive Learning of Sentence Embeddings.
\newblock In \emph{Empirical Methods in Natural Language Processing (EMNLP)}.

\bibitem[{Gu, Deng, and Su(2023)}]{gu2023dont}
Gu, Y.; Deng, X.; and Su, Y. 2023.
\newblock Don{'}t Generate, Discriminate: A Proposal for Grounding Language Models to Real-World Environments.
\newblock In Rogers, A.; Boyd-Graber, J.; and Okazaki, N., eds., \emph{Proceedings of the 61st Annual Meeting of the Association for Computational Linguistics (Volume 1: Long Papers)}, 4928--4949. Toronto, Canada: Association for Computational Linguistics.

\bibitem[{Gu et~al.(2021)Gu, Kase, Vanni, Sadler, Liang, Yan, and Su}]{gu2021beyond}
Gu, Y.; Kase, S.; Vanni, M.; Sadler, B.; Liang, P.; Yan, X.; and Su, Y. 2021.
\newblock Beyond IID: three levels of generalization for question answering on knowledge bases.
\newblock In \emph{Proceedings of the Web Conference 2021}, 3477--3488.

\bibitem[{Gu and Su(2022)}]{gu2022arcaneqa}
Gu, Y.; and Su, Y. 2022.
\newblock ArcaneQA: Dynamic Program Induction and Contextualized Encoding for Knowledge Base Question Answering.
\newblock In \emph{Proceedings of the 29th International Conference on Computational Linguistics}, 1718--1731.

\bibitem[{Izacard et~al.(2022)Izacard, Lewis, Lomeli, Hosseini, Petroni, Schick, Dwivedi-Yu, Joulin, Riedel, and Grave}]{izacard2022few}
Izacard, G.; Lewis, P.; Lomeli, M.; Hosseini, L.; Petroni, F.; Schick, T.; Dwivedi-Yu, J.; Joulin, A.; Riedel, S.; and Grave, E. 2022.
\newblock Few-shot learning with retrieval augmented language models.
\newblock \emph{arXiv preprint arXiv:2208.03299}.

\bibitem[{Johnson, Douze, and J{\'e}gou(2019)}]{johnson2019billion}
Johnson, J.; Douze, M.; and J{\'e}gou, H. 2019.
\newblock Billion-scale similarity search with {GPUs}.
\newblock \emph{IEEE Transactions on Big Data}, 7(3): 535--547.

\bibitem[{Lan and Jiang(2020)}]{lan2020query}
Lan, Y.; and Jiang, J. 2020.
\newblock Query Graph Generation for Answering Multi-hop Complex Questions from Knowledge Bases.
\newblock In \emph{Proceedings of the 58th Annual Meeting of the Association for Computational Linguistics}, 969--974.

\bibitem[{Li et~al.(2023{\natexlab{a}})Li, Sun, Tang, Yan, Wu, Huang, and Qiu}]{li2023codeie}
Li, P.; Sun, T.; Tang, Q.; Yan, H.; Wu, Y.; Huang, X.; and Qiu, X. 2023{\natexlab{a}}.
\newblock CodeIE: Large Code Generation Models are Better Few-Shot Information Extractors.
\newblock In Rogers, A.; Boyd{-}Graber, J.~L.; and Okazaki, N., eds., \emph{Proceedings of the 61st Annual Meeting of the Association for Computational Linguistics (Volume 1: Long Papers), {ACL} 2023, Toronto, Canada, July 9-14, 2023}, 15339--15353. Association for Computational Linguistics.

\bibitem[{Li et~al.(2023{\natexlab{b}})Li, Ma, Zhuang, Gu, Su, and Chen}]{li2023shot}
Li, T.; Ma, X.; Zhuang, A.; Gu, Y.; Su, Y.; and Chen, W. 2023{\natexlab{b}}.
\newblock Few-shot In-context Learning on Knowledge Base Question Answering.
\newblock In \emph{Proceedings of the 61st Annual Meeting of the Association for Computational Linguistics (Volume 1: Long Papers)}, 6966--6980. Toronto, Canada: Association for Computational Linguistics.

\bibitem[{Li et~al.(2023{\natexlab{c}})Li, Zhao, Chia, Ding, Bing, Joty, and Poria}]{li2023chain}
Li, X.; Zhao, R.; Chia, Y.~K.; Ding, B.; Bing, L.; Joty, S.; and Poria, S. 2023{\natexlab{c}}.
\newblock Chain of Knowledge: A Framework for Grounding Large Language Models with Structured Knowledge Bases.
\newblock \emph{arXiv preprint arXiv:2305.13269}.

\bibitem[{Mallen et~al.(2023)Mallen, Asai, Zhong, Das, Khashabi, and Hajishirzi}]{mallen2023trust}
Mallen, A.; Asai, A.; Zhong, V.; Das, R.; Khashabi, D.; and Hajishirzi, H. 2023.
\newblock When Not to Trust Language Models: Investigating Effectiveness of Parametric and Non-Parametric Memories.
\newblock In \emph{Proceedings of the 61st Annual Meeting of the Association for Computational Linguistics (Volume 1: Long Papers)}, 9802--9822. Toronto, Canada: Association for Computational Linguistics.

\bibitem[{McCarthy(1960)}]{mccarthy1960recursive}
McCarthy, J. 1960.
\newblock Recursive functions of symbolic expressions and their computation by machine, part I.
\newblock \emph{Communications of the ACM}, 3(4): 184--195.

\bibitem[{Shu et~al.(2022)Shu, Yu, Li, Karlsson, Ma, Qu, and Lin}]{shu2022tiara}
Shu, Y.; Yu, Z.; Li, Y.; Karlsson, B.; Ma, T.; Qu, Y.; and Lin, C.-Y. 2022.
\newblock TIARA: Multi-grained Retrieval for Robust Question Answering over Large Knowledge Base.
\newblock In \emph{Proceedings of the 2022 Conference on Empirical Methods in Natural Language Processing}, 8108--8121.

\bibitem[{Su et~al.(2016)Su, Sun, Sadler, Srivatsa, G{\"u}r, Yan, and Yan}]{su2016generating}
Su, Y.; Sun, H.; Sadler, B.; Srivatsa, M.; G{\"u}r, I.; Yan, Z.; and Yan, X. 2016.
\newblock On generating characteristic-rich question sets for qa evaluation.
\newblock In \emph{Proceedings of the 2016 Conference on Empirical Methods in Natural Language Processing}, 562--572.

\bibitem[{Sun, Bedrax-Weiss, and Cohen(2019)}]{sun2019pullnet}
Sun, H.; Bedrax-Weiss, T.; and Cohen, W. 2019.
\newblock PullNet: Open Domain Question Answering with Iterative Retrieval on Knowledge Bases and Text.
\newblock In \emph{Proceedings of the 2019 Conference on Empirical Methods in Natural Language Processing and the 9th International Joint Conference on Natural Language Processing (EMNLP-IJCNLP)}, 2380--2390.

\bibitem[{Sun et~al.(2020)Sun, Zhang, Cheng, and Qu}]{sun2020sparqa}
Sun, Y.; Zhang, L.; Cheng, G.; and Qu, Y. 2020.
\newblock SPARQA: skeleton-based semantic parsing for complex questions over knowledge bases.
\newblock In \emph{Proceedings of the AAAI conference on artificial intelligence}, 8952--8959.

\bibitem[{Tan et~al.(2023)Tan, Chen, Shao, and Chen}]{tan2023make}
Tan, C.; Chen, Y.; Shao, W.; and Chen, W. 2023.
\newblock Make a Choice! Knowledge Base Question Answering with In-Context Learning.
\newblock \emph{arXiv preprint arXiv:2305.13972}.

\bibitem[{Wang, Li, and Ji(2022)}]{wang2022code4struct}
Wang, X.; Li, S.; and Ji, H. 2022.
\newblock Code4struct: Code generation for few-shot structured prediction from natural language.
\newblock \emph{arXiv preprint arXiv:2210.12810}.

\bibitem[{Wang et~al.(2022)Wang, Wei, Schuurmans, Le, Chi, Narang, Chowdhery, and Zhou}]{wang2022self}
Wang, X.; Wei, J.; Schuurmans, D.; Le, Q.~V.; Chi, E.~H.; Narang, S.; Chowdhery, A.; and Zhou, D. 2022.
\newblock Self-Consistency Improves Chain of Thought Reasoning in Language Models.
\newblock In \emph{The Eleventh International Conference on Learning Representations}.

\bibitem[{Wei et~al.(2022)Wei, Wang, Schuurmans, Bosma, Xia, Chi, Le, Zhou et~al.}]{wei2022chain}
Wei, J.; Wang, X.; Schuurmans, D.; Bosma, M.; Xia, F.; Chi, E.; Le, Q.~V.; Zhou, D.; et~al. 2022.
\newblock Chain-of-thought prompting elicits reasoning in large language models.
\newblock \emph{Advances in Neural Information Processing Systems}, 35: 24824--24837.

\bibitem[{Ye et~al.(2022)Ye, Yavuz, Hashimoto, Zhou, and Xiong}]{ye2022rng}
Ye, X.; Yavuz, S.; Hashimoto, K.; Zhou, Y.; and Xiong, C. 2022.
\newblock RNG-KBQA: Generation Augmented Iterative Ranking for Knowledge Base Question Answering.
\newblock In \emph{Proceedings of the 60th Annual Meeting of the Association for Computational Linguistics (Volume 1: Long Papers)}, 6032--6043.

\bibitem[{Yih et~al.(2016)Yih, Richardson, Meek, Chang, and Suh}]{yih2016value}
Yih, W.-t.; Richardson, M.; Meek, C.; Chang, M.-W.; and Suh, J. 2016.
\newblock The value of semantic parse labeling for knowledge base question answering.
\newblock In \emph{Proceedings of the 54th Annual Meeting of the Association for Computational Linguistics (Volume 2: Short Papers)}, 201--206.

\bibitem[{Yu et~al.(2022)Yu, Zhang, Ng, Zhu, Li, Wang, Hu, Wang, Wang, and Xiang}]{yu2022decaf}
Yu, D.; Zhang, S.; Ng, P.; Zhu, H.; Li, A.~H.; Wang, J.; Hu, Y.; Wang, W.~Y.; Wang, Z.; and Xiang, B. 2022.
\newblock DecAF: Joint Decoding of Answers and Logical Forms for Question Answering over Knowledge Bases.
\newblock In \emph{The Eleventh International Conference on Learning Representations}.

\bibitem[{Zhou et~al.(2022)Zhou, Sch{\"a}rli, Hou, Wei, Scales, Wang, Schuurmans, Cui, Bousquet, Le et~al.}]{zhou2022least}
Zhou, D.; Sch{\"a}rli, N.; Hou, L.; Wei, J.; Scales, N.; Wang, X.; Schuurmans, D.; Cui, C.; Bousquet, O.; Le, Q.~V.; et~al. 2022.
\newblock Least-to-Most Prompting Enables Complex Reasoning in Large Language Models.
\newblock In \emph{The Eleventh International Conference on Learning Representations}.

\end{thebibliography}

\newpage
\clearpage
\appendix
\section{Appendix}

\subsection{Adaptive modification based on S-Expression} \label{appx_sec:meta-functions}
Compared to the original syntax of S-Expression, we have made the following adaptive modifications:
\begin{itemize}
    \item {\bf Omit the R function}. The {\bf R} function refers to the direction of the relation, but the direction can be determined by relation matching.
    \item {\bf Modifying the JOIN function}. The {\bf JOIN} function in the original syntax can be called as $(E',E'') = (${\bf JOIN}$\ r_1\ r_2)$, where $r_1, r_2 \in \mathcal{R}$, $E'$ represents the header entity set of $r_1$ and $E''$ represents the tailed entity set of $r_2$. However, this call form is unnecessary and can be replaced by other combinations of function calls. Thus, we removed this calling form to ensure that {\bf JOIN}, like other functions, only returns a single entity set.
    \item {\bf Add the {\bf START} and {\bf STOP} functions}. These two functions are for better labeling the start and end of expression generation.
\end{itemize}

\subsection{More Details on Experiment}
\paragraph{Evaluation Dataset}
\begin{itemize}
    \item {\bf WebQSP} is an I.I.D generalized dataset, and the entities and relations in the test set of problems have all appeared in the training set. The problems mainly contain 1-hop or 2-hop relations, and there are entity constraints and aggregation constraints in some questions.
    \item {\bf GrailQA} is a dataset containing three levels of generalization, including I.I.D, compositional, and zero-shot. Compared to other datasets, its main feature lies in the inclusion of the zero-shot generalization questions, which involve unseen schema items or even domains in the training set.
    \item {\bf GraphQ} is a compositional generalized dataset. It further enhances WebQSP in terms of syntactic structure and expressive diversity, contains numerical constraints other than time, and also suffers from query counting.
\end{itemize}
\begin{table}[ht]
    \centering
    \begin{tabular}{ccccc}
        \toprule
        \multirow{2}{*}{\bf Dataset} & {\bf Generalization} & \multirow{2}{*}{\bf Train} & \multirow{2}{*}{\bf Dev} & \multirow{2}{*}{\bf Test} \\
        & {\bf Level} & & & \\
        \midrule
        WebQSP & I.I.D & 3,098 & - & 1,639 \\
        \midrule
        GraphQ & Compositional & 2,381 & - & 2,395 \\
        \midrule
        \multirow{3}{*}{GrailQA} & I.I.D \& & \multirow{3}{*}{44,337} & \multirow{3}{*}{6,763} & \multirow{3}{*}{13,231*} \\
        & Compositional \& &  &  &  \\
        & Zero-shot &  &  &  \\
        \bottomrule
    \end{tabular}
    \caption{Evaluation dataset statistics. *: Performance on GrailQA is all reported on the local dev dataset in this paper and does not involve the test set.}
    \label{appx_tab:dataset}
\end{table}

\paragraph{Variant of KB-Coder}
We describe the majority vote strategy and similarity sampling used in the variant of KB-Coder. For the majority vote strategy, which is used in KB-Coder(6) and KB-Coder(6)-R, we use the \texttt{n} field in the API provided by OpenAI to make \texttt{gpt-3.5-turbo} generate 6 independent function call sequences for the same input. Subsequently, when computing FER, we label a particular test question as ``formatting error'' if and only if the formatting of all generated function call sequences is wrong. When computing the F1 Score, we get the answer for each generated function call sequence followed by the method described in the main text. Based on the self-consistency in LLM, we consider the answer with the most occurrences as the predicted answer.

For the similar sampling of demo examples, which is used in KB-Coder(1)-R and KB-Coder(6)-R, we also use \texttt{sup-simcse-bert-base-uncased} to obtain the representation of the questions in the training and test dataset. Then, we use \texttt{Faiss} to build the index for all questions in the training dataset and retrieve the most similar $K$ questions for each test question. Then the selected questions and their S-Expression are converted into the demo examples.

\subsection{Discussion}

\subsubsection{In the age of LLM, is KBQA still necessary?}
We believe that KBQA is necessary. In the initial experiments, we randomly sampled 100 questions from WebQSP and tried to let the LLM give the answers directly. After the human judgment of the answers, \textit{Text-Davinci-003} obtains 67\% Hit@1. However, we also found that the LLM has some problems that cannot be solved in the short term: (1) The LLMs tend to give the most well-known answer instead of all correct answers; (2) the LLMs tend to give a wider range of answers that are not wrong rather than an exact answer; (3) In addition to answering incorrectly, the LLMs run the risk of generating things that do not exist in the real world. By generating the correct query statements, we can simultaneously ensure that the answers are comprehensive, precise, and truthful, properties that LLMs do not have when they give answers.

\subsubsection{What are the advantages of KB-Coder over others?}
The core advantages of KB-Coder over existing methods are threefold: training free, no fake risk, and providing reasoning steps. First, KB-Coder with in-context learning eliminates the need for fine-tuning, enabling rapid deployment in different domains; second, KB-Coder generates logical forms instead of directly generating answers, thus avoiding the risk of generating false knowledge directly from the mechanism; finally, KB-Coder provides complete reasoning steps instead of directly generating the whole logical form, which makes the answers more interpretable and easier to further improve in future. In Table 6, We compare KB-Coder with other published works and contemporaneous works to demonstrate the characteristics of our methods.

\begin{table}[ht]
    \centering
    \begin{tabular}{l|ccc}
        \toprule
        Advantage & DecAF & KB-BINDER & Ours\\
        \midrule
        Training-Free & \faTimes & \faCheck & \faCheck \\
        No Fake Answer & \faTimes & \faCheck & \faCheck \\
        Reasoning Step & \faTimes & \faTimes & \faCheck \\
        \bottomrule
    \end{tabular}
    \caption{Characteristics of our method compared to other methods. We used CoK plus the author's last name to distinguish between two same-name methods.}
    \label{appx_tab:advantage}
\end{table}

\subsubsection{What part of KB-Coder could be improved?}
We find that the strategy of entity linking and relation matching could greatly affect the performance of the KB-Coder.  Our strategy is to choose the first candidate item for the function call sequence that can be queried to obtain non-empty answers. To explore the upper limit of the KB-Coder, we evaluate all candidates and choose the one with the highest F1 score. In this setting, the performance of KB-Coder(6)-R on GraphQ improves from 36.6 to 50.2, illustrating the difficulty of entity linking and relation matching.

\subsection{A Complete Case of Code-Style ICL}
\begin{small}
\begin{minted}{Python}
'''
Please use the functions defined below to generate 
the expression corresponding to the question step by step.
'''
def START(entity: str):
    return entity

def JOIN(relation: str, expression: str):
    return '(JOIN {} {})'.format(relation, expression)

def AND(expression:str, sub_expression: str):
    return '(AND {} {})'.format(expression, sub_expression)

def ARG(operator: str, expression: str, relation: str):
    assert operator in ['ARGMAX', 'ARGMIN']
    return '({} {} {})'.format(operator, expression, relation)

def CMP(operator: str, relation: str, expression: str):
    assert operator in ['<', '<=', '>', '>=']
    return '({} {} {})'.format(operator, relation, expression)

def COUNT(expression: str):
    return '(COUNT {})'.format(expression)

def STOP(expression: str):
    return expression

question = 'road runner railway has what material?'
expression = START('road runner railway')
expression = JOIN('amusement_parks.roller_coaster_material.roller_coasters', expression)
expression = AND('amusement_parks.roller_coaster_material', expression)
expression = STOP(expression)

(There are 38 demo examples omitted here.)


expression = START('skytrain')
expression = JOIN('travel.travel_destination.local_transportation', expression)
expression = AND('travel.travel_destination', expression)
expression = STOP(expression)

'''
Some relations for reference are as follows:
rail.railway.part_of_network
'''

question = 'semaphore railway line is on the rail network named what?'


\end{minted}
\end{small}

\end{document}